# A Hierarchical fuzzy controller for a biped robot.

Abdallah Zaidi , Nizar Rokbani, and Adel. M. Alimi, *Senior Member IEEE*
REGIM Lab.: REsearch Groups on Intelligent Machines
National Engineering School of Sfax,
University of Sfax, BP 1173, Sfax, 3038, Tunisia
{Abdallah.zaidi, nizar.rokbani, adel.alimi}@ieee.org

*Abstract*—**In this paper the investigation is placed on the hierarchic neuro-fuzzy systems as a possible solution for biped control. An hierarchic controller for biped is presented, it includes several sub-controllers and the whole structure is generated using the adaptive Neuro-fuzzy method. The proposed hierarchic system focus on the key role that the centre of mass position plays in biped robotics, the system sub-controllers generate their outputs taken into consideration the position of that key point.**

**Keywords: biped robot; Neuro-fuzzy systems; hierarchical fuzzy systems.**

## I. INTRODUCTION

The control of a humanoid robot is a challenging task due to the hard-to stabilize. In recent years, the control of biped robots has made great progress. The control of a biped walking robot is to find a control law that allows the coordination of movements of the various members of the articulated mechanical structure [2]. This enables the movement of the robot on a ground. Different methods were developed for the control of biped robots with a focus on the stabilization of biped locomotion system [3, 4].

Biped robots are a typical case of non-linear complex system; the control of such systems is addressed using classical methodologies such as the PID and also intelligent techniques such as PSO, fuzzy sets, neural networks and neuro-fuzzy systems [4, 5, 6, 7].

The IZIMAN is a research projects that conducting in REGIM laboratory, "Research group on Intelligent Machines". The main challenge of the project is to propose an intelligent architecture and controller that are "humanly" inspired [8]. Within this project several gait generation methodologies focusing on PSO [9, 10, 11, 13], and also neuro-fuzzy was investigated [7], PSO based biped control was implemented on a small size robot [12].

In this paper the investigation is placed on the hierarchic neuro-fuzzy systems as a possible solution for biped control. An hierarchic controller for biped is presented, it includes several sub-controllers and the whole structure is generated using the adaptive Neuro fuzzy method.

This paper is organized as follows: the first section include the description of neuro-fuzzy and hierarchical systems. In the second section, the hierarchical fuzzy controller for biped robot is presented. In the third section, we present the controllers design, obtained results are presented in paragraph four. Finally, paragraph five include the conclusions and further work.

## II. NEURO-FUZZY AND HIERARCHIC FUZZY SYSTEMS

### A. Neuro-fuzzy systems

Artificial neural networks (ANN) and fuzzy inference system (FIS) are generally considered to be complementary areas of research. The combination of FIS and ANN define a neuro-fuzzy system in such a way that the parameters of FIS are determined by using the neural network learning algorithms [14].

The neuro-fuzzy system uses the linguistic knowledge of fuzzy inference system and the learning capability of neural network. To describe the architecture a neuro-fuzzy system, consider Fig1. For simplicity, we assume that our fuzzy system has two inputs and one output. Furthermore, we assume that the defuzzification of the variables is a linear combination of the first order of input variables [14, 15].

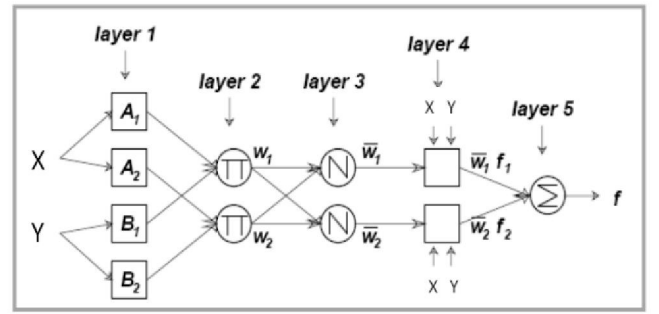

Fig1. neuro-fuzzy system architecture

### B. Hierarchical fuzzy systems

Hierarchical fuzzy systems have been created to address one of the major problems of standard fuzzy systems which is essentially the impact of the number of input variables on the number of rules of the system [16, 17]. The use of a hierarchic structure allowed designing a set of sub-controllers with limited number of variables for each one, see Figure 2.

Complex structures including a hierarchic organization are the typical target of the HFLC design concept. In bio-inspired systems, such as the human locomotion system, a hierarchic design could be directly inspired from the system. For such systems the HLFC have a build in justification.

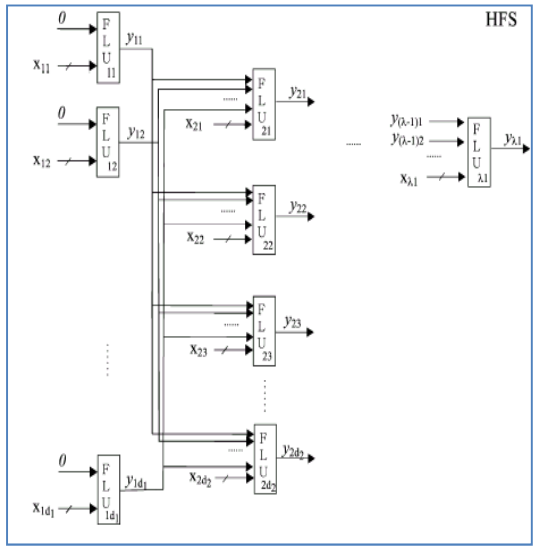

Fig2. Hierarchical fuzzy systems architecture as in [20]

### III. AN HIERARCHICAL FUZZY SYSTEM FOR BIPED ROBOT CONTROL

A planar biped model is used, the model includes a knee rotation, a hip rotation, and the remaining of the body is represented by a punctual mass with a motion, see figure3, such a model is close to the "Bipsim" proposed in [22].

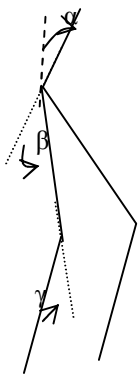

Fig4. Structure of the biped robot

The training Data-set needed to generate the controllers using the Adaptive Neuro-Fuzzy System method is obtained from a classical dynamic model similar to [18]. Separated Data-sets of 10, 30, 40, 60 and 120 samples are created in order to evaluate the impact of the training set size on the quality of the controllers. Test data set relative to training data sets are separately managed.

The overall architectural design is given in figure 4, where a set of neuro-fuzzy controllers integrated into a Hierarchical structure centered on the COM position of the biped robot. The system generates the needed joints rotations for a given COM reference position, a supervision system is then used to compute the estimated position of the COM.

Due to the symmetry of the walking system, the design of the left and right legs are the same, a leg include four controllers. The left leg controllers are HFLC1, HFLC3, HFLC5 and HFLC7, the right leg controllers are HFLC2, HFLC4, HFLC6 and HFLC 8. Since the controllers design is similar only the left leg controllers are described in this paper. The overall design appears in figure5, a brief description of the sub-controllers is given bellow:

- For HFL1, the controller inputs are the vector (x0, y0) coordinates of the center of mass and the angle angel (beta_left), the output is the same leg angle (gamma_left).
- For HFL3, the controller inputs are the vector (x0, y0) coordinates of the center of mass and the angle angel (gamma_left), the output is the vector (xcl,ycl) coordinates of left ankle.
- For HFL5, the controller inputs are the vector (x0, y0) coordinates of the center of mass the vector (xcl,ycl) coordinates of left ankle. The output is the angel (beta_left).

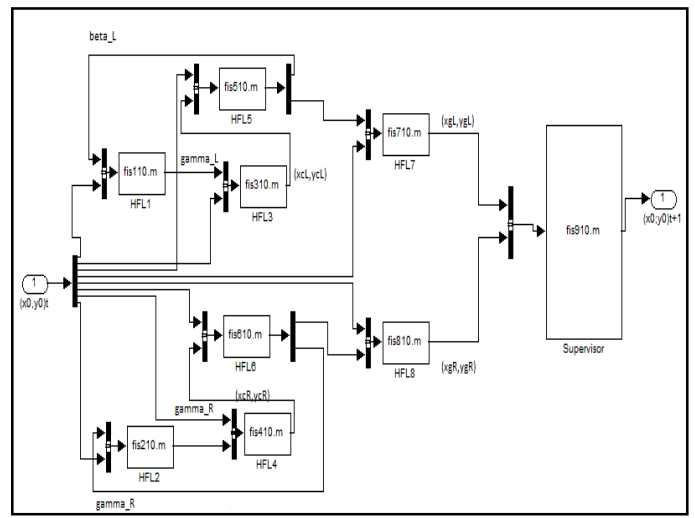

Fig5. Hierarchic fuzzy controller for biped robot

### IV. RESULTS

#### A. Controllers Responses

In this paragraph we presented the controllers response surface, representing controller output in accordance with its inputs. The controller's behavior corresponding to the left leg and using a training data set of 10 samples appears in figure 6.

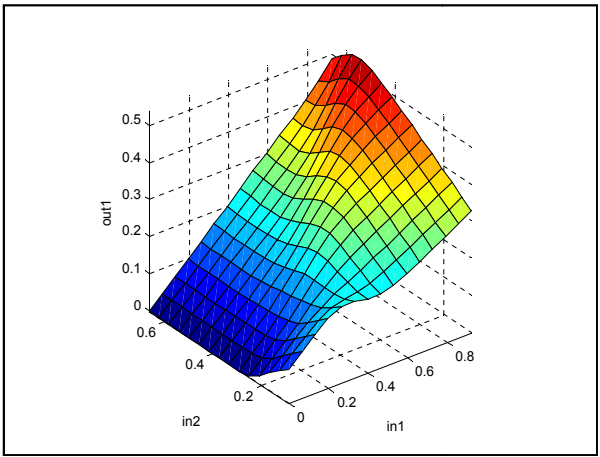

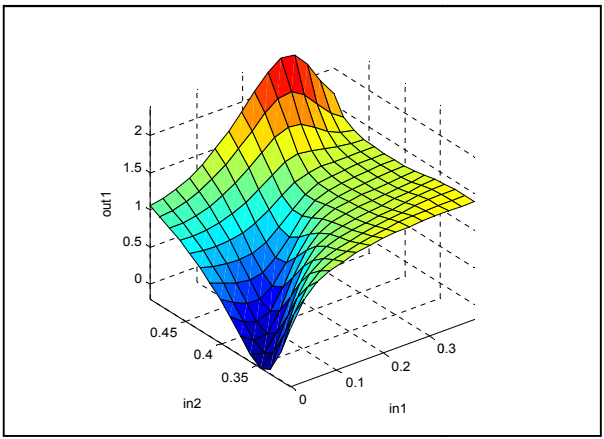

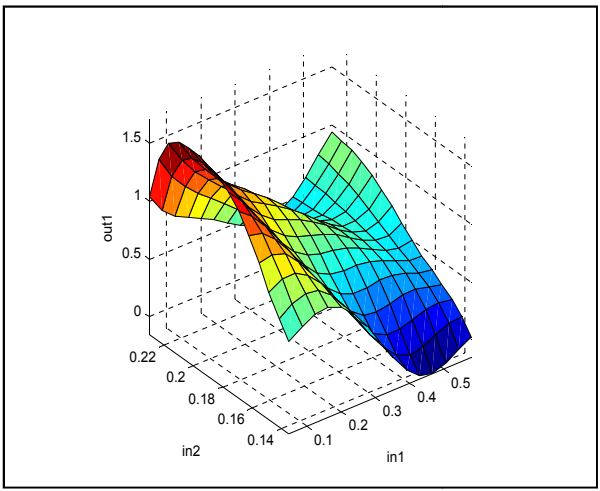

Fig6. Controllers Design, from top to bottom HFLC1, HFLC2 and HFLC3 controllers responses.

## B. Effect of the training data size

The controllers are generated using a set of training Data including respectively 10, 30, 40 and 120 samples; then they are evaluated using a set of test Data. The cumulative scare error for all test data is computed, it is used as an evaluation criteria of the HFLCs. Results are presented in figures 7, 8 and 9.

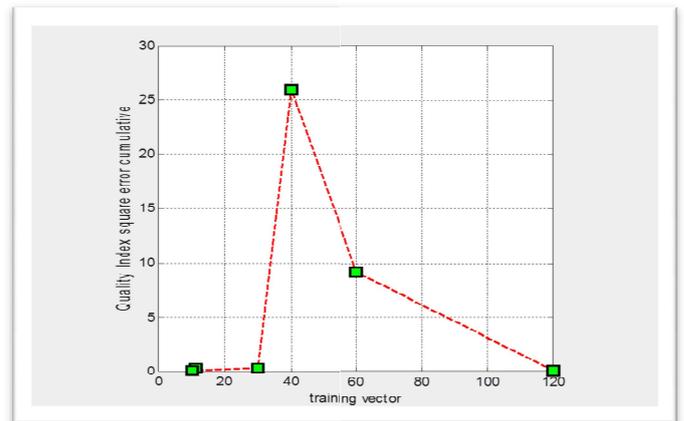

Fig7. representation of the error of the proposed HFLC1

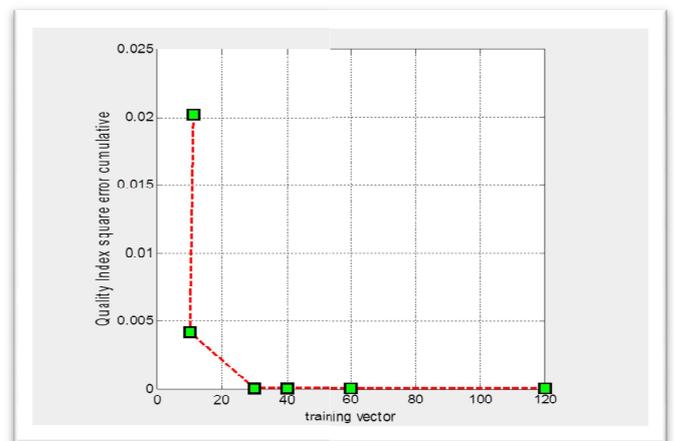

Fig8. Representation of the error of HFLC3

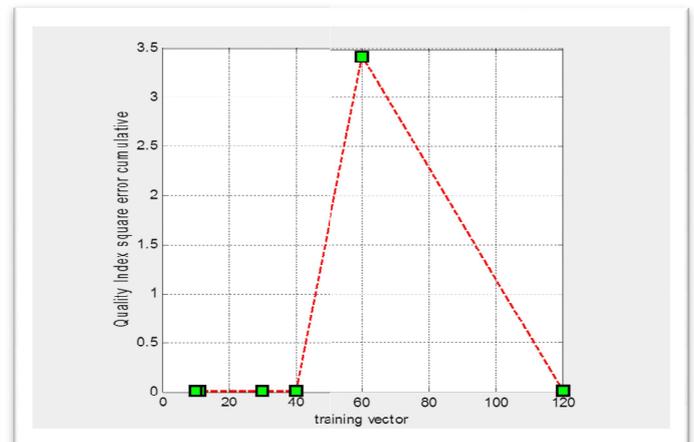

Fig9. Representation of the error of HFLC5

The results show that the best training set size for all controllers is made of 30 samples. HFLC1 had an error of (25) for a training set of (40) samples, a similar result is observed in HFLC5 when trained with (60) samples its error is about (3.5).

For a training set of 30 samples the error is about (0.001) for HFLC1, (0.00001) for HFLC3 and (0.0001) in the case of HFLC5.

## V. DISCUSSION AND FURTHER WORK

In this paper, we have proposed a hierarchical fuzzy controller of the biped robot. The generation methodology as well as the test procedure is explained. An investigation on the effect of the number of training samples is also conducted and allowed to fix the optimal size to (30) experimentally.

The paper detailed only the generation and test of the left-leg since the right leg has a symmetric Behavior. The obtained results are interesting with a limited error for the main controllers of the left leg.

A comparative test between the full controller and a classical one such in [18] will be developed soon.

## ACKNOWLEDGMENT

The authors would like to acknowledge the financial support of this work by grants from General Direction of Scientific Research (DGRST), Tunisia, under the ARUB program.